%% file: camera_ready_paper.tex
\documentclass[10pt,twocolumn,letterpaper]{article}

\usepackage{3dv}
\usepackage{times}
\usepackage{epsfig}
\usepackage{graphicx}
\usepackage{amsmath}
\usepackage{amssymb}

\usepackage{booktabs}
\usepackage{multirow}
\usepackage[utf8]{inputenc}
\usepackage{subcaption}
\usepackage[numbers,sort,compress]{natbib}
\usepackage{balance}

\usepackage[pagebackref=true,breaklinks=true,colorlinks,bookmarks=false]{hyperref}

\threedvfinalcopy 

\def\threedvPaperID{148} 

\newcommand\blfootnote[1]{%
  \begingroup
  \renewcommand\thefootnote{}\footnote{#1}%
  \addtocounter{footnote}{-1}%
  \endgroup
}

\ifthreedvfinal\fi
\begin{document}

\title{Human Body Measurement Estimation with Adversarial Augmentation}


\author{
Nataniel Ruiz$^{2\dag}$ 
\qquad 
Miriam Bellver$^1$
\qquad
Timo Bolkart$^1$
\qquad 
Ambuj Arora$^1$\\
\qquad
Ming C. Lin$^1$
\qquad 
Javier Romero$^{3\dag}$
\qquad 
Raja Bala$^1$
\\
\textrm{$^1$Amazon 
\qquad $^2$Boston University 
\qquad $^3$Reality Labs Research} 
\\
{\tt\small nruiz9@bu.edu}
\quad
{\tt\small \{mbellver, timbolka,  ambarora, minglinz, rajabl\}@amazon.com}
}

\maketitle

\blfootnote{\dag This research was performed while NR and JR were at Amazon.}
\begin{abstract}
  We present a Body Measurement network~(BMnet) for estimating 3D anthropomorphic measurements of the human body shape from silhouette images. Training of BMnet is performed on data from real human subjects, and augmented with a novel adversarial body simulator~(ABS) that finds and synthesizes challenging body shapes. ABS is based on the skinned multiperson linear~(SMPL) body model, and aims to maximize BMnet measurement prediction error with respect to latent SMPL shape parameters. ABS is fully differentiable with respect to these parameters, and trained end-to-end via backpropagation with BMnet in the loop. Experiments show that ABS effectively discovers adversarial examples, such as bodies with extreme body mass indices (BMI), consistent with the rarity of extreme-BMI bodies in BMnet's training set. Thus ABS is able to reveal gaps in training data and potential failures in predicting under-represented body shapes. Results show that training BMnet with ABS improves measurement prediction accuracy on real bodies by up to {\bf 10\%}, when compared to no augmentation or random body shape sampling. Furthermore, our method significantly outperforms SOTA measurement estimation methods by as much as {\bf 3x}. Finally, we release \textit{BodyM}, the first challenging, large-scale dataset of photo silhouettes and body measurements of real human subjects, to further promote research in this area. Project website: \url{https://adversarialbodysim.github.io}.
\end{abstract}
\vspace{-10pt}


\begin{figure*}[t]
\centering
\begin{subfigure}{0.7\columnwidth}
\includegraphics[width=\columnwidth]{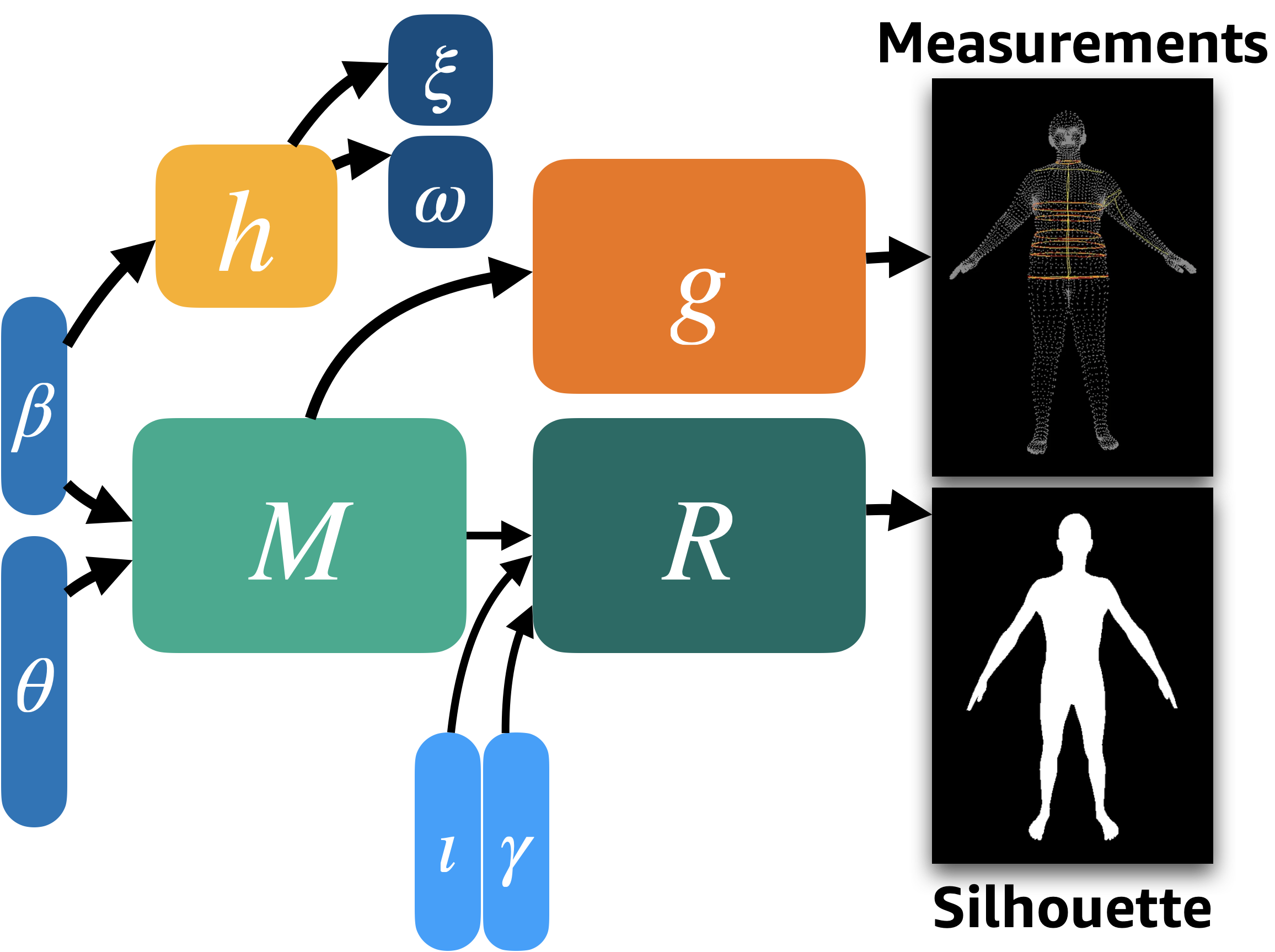}
\caption{}
\label{fig:renderer}
\end{subfigure}%
\hspace{10pt}
\begin{subfigure}{0.7\columnwidth}
\includegraphics[width=\columnwidth]{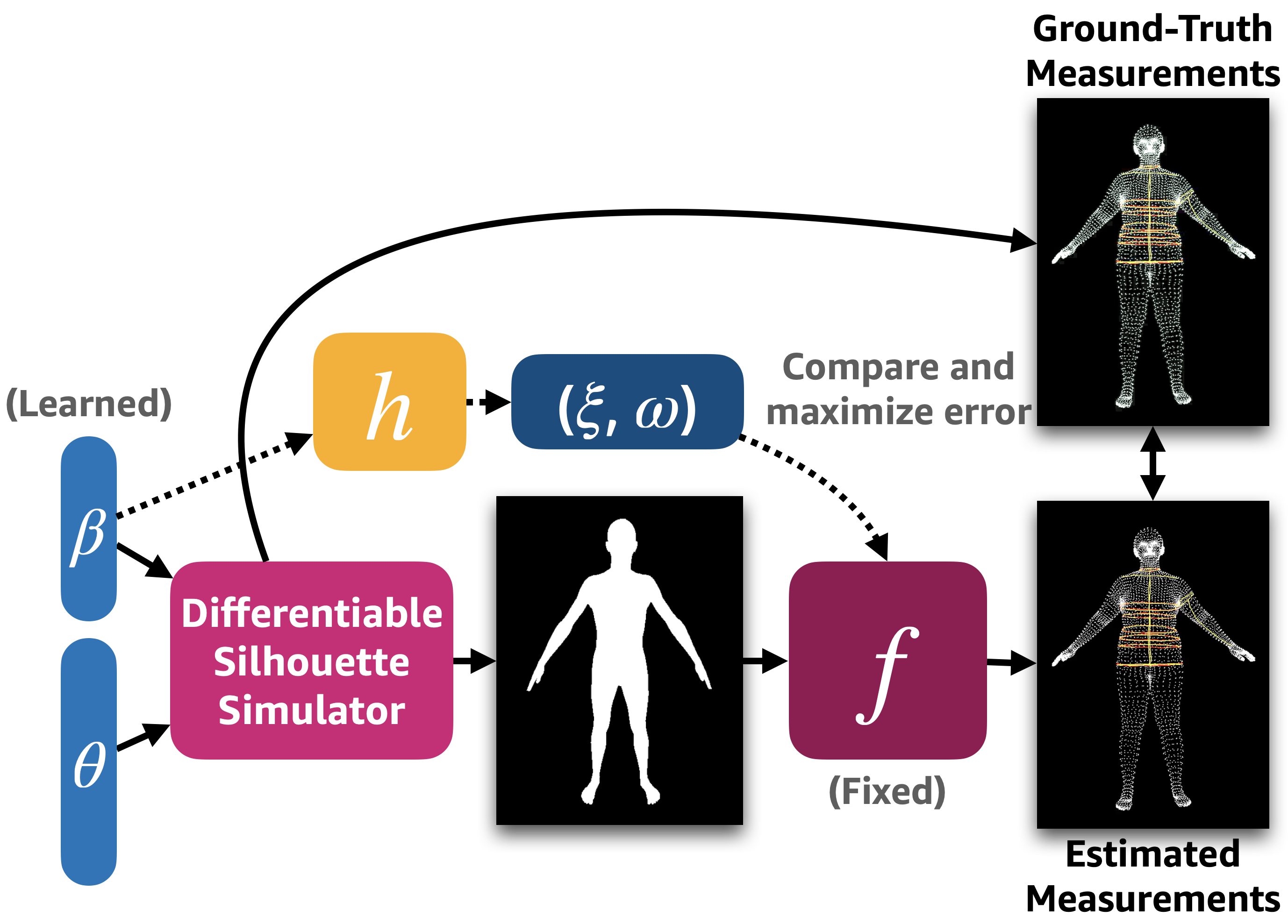}
\caption{}
\label{fig:adv_shape_opt}
\end{subfigure}%
\caption{{\textbf{(a)} \textbf{Differentiable silhouette simulator:}} SMPL model M generates a body mesh from shape and pose parameters $\beta$ and $\theta$, which is passed to silhouette renderer $R$ (parameterized by lighting $\iota$ and camera $\gamma$), and measurement extractor $g$. Regressor $h$ generates height $\xi$ and weight $\omega$ from $\beta$. \textbf{(b)} \textbf{Adversarial shape optimization:} The simulator renders silhouettes that are passed to BMnet (\textit{f}) along with height $\xi$ and weight $\omega$ to obtain measurement estimates, which are compared to ground truth measurements (also generated by the simulator). The error is maximized with respect to shape $\beta$ under fixed pose $\theta$. }
\vspace{-10pt}
\end{figure*}

\section{Introduction}
\label{sec:intro}
Reconstruction of the 3D human body shape from images is an important problem in computer vision which has received much attention in the last few years \cite{Wang_2021_survey,SMPL:2015,Saito_PIFU_HD_CVPR2020,Biggs2020,hmrKanazawa17,HoloPose2019,Omran2018NBF,Rockwell2020,Kocabas_PARE_2021,Kocabas_SPEC_2021,Liang_ICCV2019,ExPose:2020,PIXIE:2021,FrankMocap2021,Choi_ECCV20,kolotouros2019convolutional,Fan_2021_ICCV,Wan_2021_ICCV,Zhang_2021_ICCV,li2021everybody,choi20213dcrowdnet,pavlakos2020human,wei2022capturing,yuan2021glamr,Zanfir_2021_CVPR,Shapy:CVPR:2022}. However, 3D shape is not directly usable for applications where anthropomorphic body measurements are required. In healthcare, for example, measurements such as waist girth are a key indicator of body fat; while in the fashion industry, metric body measurements enable size recommendations and made-to-measure garments. Surprisingly, much less work has been published on directly estimating body measurements from images. This is the problem that we address in this paper. Note that body measurements can be viewed as a compact yet rich descriptor for 3D body shape. Indeed, previous work has shown that it is possible to accurately map a few body measurements to a 3D body mesh in a reference pose ~\cite{sengupta2021probabilistic, pujades2019virtual}. 

Most existing body reconstruction methods do not incorporate knowledge of camera intrinsics or scale, and thus cannot guarantee metric accuracy~(i.e. the distance between two points on the recovered mesh may not correspond to physical distances on a person's body) ~\cite{bodytalk2016,kolotouros2019learning,sengupta2020synthetic}. Furthermore, since these approaches have only been trained to generate a posed 3D avatar of a human, the body measurements have to be derived from the predicted mesh, which can limit resolution and accuracy. Finally, acquiring physical body measurements at scale is costly and time-consuming; hence, there is a dearth of training datasets pairing images with measurements of real humans. To circumvent this challenge, previous efforts have used synthetic data for training~\cite{dibra2016hs,Smith2019TowardsA3}, and evaluated on very small numbers (2-4) of human subjects~\cite{Boisvert_2013,dibra2016hs}.

We present a method to predict body measurements from images that alleviates these shortcomings. We train a convolutional body measurement network~(\textit{BMnet}) to directly predict measurements from two silhouette images of a person's body. Silhouettes effectively convey body shape information, while preserving user privacy. To resolve scale ambiguity, we include height and weight as additional inputs to BMnet. We introduce a novel adversarial body simulator~(\textit{ABS}) that automatically discovers and synthesizes body shapes for which BMnet produces large prediction errors. ABS is fully differentiable with BMnet in-the-loop. It uncovers weaknesses in the model and gaps in the training data. For example, body shapes returned by ABS tend to be of predominantly high body-mass-index~(BMI), consistent with the fact that these shapes are under-represented in training. Fine-tuning BMnet with samples generated by ABS improves accuracy (up to {\em 10\%}) and robustness on real data, achieving state-of-art results. 
To train and evaluate BMnet, we introduce a new dataset, \textit{BodyM}, comprising full-body silhouette images of 2,505 subjects in frontal and lateral poses, accompanied by height, weight, and 14 body measurements derived from 3D scans. To our knowledge, this is the first dataset that pairs photo silhouettes and body measurements for real humans at such a scale.

The main contributions of this work are: 
\begin{itemize}
    \vspace*{-0.5em}
    \item \textit{BMnet}: A deep CNN to directly regress {\em physical body measurements} from 2 silhouettes, height and weight;
    \vspace*{-0.5em}
    \item \textit{ABS}: A novel {\em differentiable simulator} for generating {\em adversarial body shapes} with BMnet in-the-loop, uncovering training gaps and improving BMnet performance on real data (up to {\bf 3x});
    \vspace*{-0.5em}
    \item \textit{BodyM}: A new {\em dataset for body measurement estimation} comprising silhouettes, height, weight and 14 physical body measurements for $2,505$ humans, publicly available for research purposes \footnote{\url{https://adversarialbodysim.github.io}}.
    \vspace*{-0.2em}
\end{itemize}


\section{Related Work}
\label{sec:relwork}

\textbf{Body reconstruction from RGB images:}
The literature on recovering 3D human representations from RGB images is vast; see \cite{tian_HMRsurvey_2021} and \cite{Wang_2021_survey}  for excellent surveys. Techniques fall broadly into two categories. Parametric methods characterize the human body in terms of a parametric model such as SMPL\{-X\}~\cite{SMPL:2015,SMPL-X:2019}, Adam~\cite{Joo_2018_CVPR}, SCAPE~\cite{Anguelov_SCAPE_2005}, STAR~\cite{STAR:2020}, or GHUM~\cite{Xu_2020_CVPR}. Model parameters defining body pose and shape are then estimated from images via direct optimization ~\cite{bogo2016keep,SMPL-X:2019,Xiang2019monocular,Zanfir_2021_CVPR}, regression with deep networks~\cite{Shapy:CVPR:2022,Biggs2020,hmrKanazawa17,HoloPose2019,Omran2018NBF,Rockwell2020,Kocabas_PARE_2021,Kocabas_SPEC_2021,Liang_ICCV2019,ExPose:2020,FrankMocap2021,zanfir2021thundr}, or a combination of the two~\cite{kolotouros2019learning}. In contrast, non-parametric methods directly regress a 3D body representation from images using graph convolutional neural networks~\cite{Choi_ECCV20,kolotouros2019convolutional}, transformers~\cite{Lin_2021_METRO}, combinations of both~\cite{Lin_2021_ICCV}, intermediate representations such as 1D heatmaps ~\cite{Moon_2020_ECCV_I2L-MeshNet} or 2D depth maps ~\cite{Smith_2019_ICCV}, or with implicit functions~\cite{Corona2021_SMPLicit,Saito_PIFU_HD_CVPR2020}. Recently, there have been successful explorations on probabilistic approaches for shape and pose estimation~\cite{kolotouros2021prohmr,sengupta2021probabilistic,Sengupta_2021_ICCV,sengupta2021probabilistic_cvpr}.

\textbf{Body reconstruction from silhouettes:}
Methods have been proposed to predict 3D body model parameters from binary human silhouette images~\cite{4270338,4409005,pavlakos2018learning,sengupta2020synthetic,Dibra_2017_CVPR}. Our approach is similar in flavor, but addresses a different task of predicting physical body measurements from silhouettes. Our constrained pose setting, height and weight inputs, and adversarial training scheme enable measurement prediction with state-of-art metric accuracy.

\textbf{Body measurement estimation:}
Dibra et al.~\cite{dibra2016hs} reported the first attempt at using a CNN to recover a 3D body mesh and anthropomorphic measurements from silhouettes. The silhouettes are generated synthetically by rendering 3D meshes from the CAESAR~(Civilian American and European Surface Anthropometry Resource) dataset~\cite{CAESAR} onto frontal and side views, and body measurements are derived as geodesic distances on 3D meshes. In contrast, our approach is trained on data from both real and synthetic humans, directly regresses measurements, and employs adversarial training for improved performance. Our approach is most closely related to the works of \cite{Smith2019TowardsA3} and  \cite{yan2020silh}. Yan et al.~\cite{yan2020silh} use their BodyFit dataset to train a CNN to predict measurements from silhouette pairs. Smith et al.~\cite{Smith2019TowardsA3} proposed a multitask CNN to estimate body measurements, body  mesh, and 3D pose from height, weight, two silhouette images and segmentation confidence maps. For training, they generate synthetic body shapes by sampling the SMPL shape space with multivariate Gaussian shape distributions and stochastic perturbations of body shapes from CAESAR. In contrast to both these methods, our approach seeks adversarial samples in the low performance regime of BMnet, enabling automatic discovery and mitigation of weaknesses in dataset and network in a principled manner.

\textbf{Synthesis for training:}
With advances in simulation quality and realism, it has become increasingly common to train deep neural networks using synthetic data~\cite{virtualkitti,synthia,playing_for_data,dosovitskiy2017carla,Liang_ICCV2019}. Recently, there have been attempts at learning to adapt distributions of generated synthetic data to improve model training~\cite{louppe2019adversarial,ruiz2018learning,Ganin2018SynthesizingPF,Beery_2020_WACV,andrychowicz2020learning,shen_NIPS2021,yao2020simulating,sampling_strategies,autosimulate}. These approaches focus on approximating a distribution that is either similar to the natural test distribution or that minimizes prediction error. Another flavor of approaches probes the weaknesses of machine learning models using synthetic data~\cite{pinto2008establishing,mayer2016large,johnson2017clevr,kortylewski2018empirically,kortylewski2019analyzing,shu_ICRA2021,ruiz2020morphgan}. The works of \cite{alcorn2019strike,zeng2019adversarial,shu2020identifying} generate robust synthetic training data for object recognition and visual-question-answering by varying scene parameters such as pose and lighting, while preserving object characteristics. Shen et al. \cite{shen2020driving} tackle vehicle self-driving by introducing adversarial camera corruptions in training. In our work, we explore the impact of varying interpretable parameters that directly control human body shape.

\textbf{Adversarial techniques:}
We take inspiration from the literature on adversarial attacks of neural networks~\cite{szegedy2013intriguing,papernot2017practical,carlini2017towards,explaining_adv} and draw from ideas for improving network robustness by training on images that have undergone white-box adversarial attacks~\cite{madry2018towards}. The main difference lies in the search space: previous works search the image space while we search the interpretable latent shape space of the body model. The works by \cite{qiu2020semanticadv,ruiz2021simulated} find synthetic adversarial samples for faces using either a GAN or a face simulator. They are successful in finding interpretable attributes leading to false predictions; however, they do not incorporate this knowledge in training to improve predictions on real examples. 
In our work, we both discover adversarial samples and use them in training to improve body measurement estimation. Different from previous methods, we find adversarial bodies by searching the latent space of a body simulator comprising a pipeline of differentiable submodules, namely: a 3D body shape model, body measurement estimation network, height and weight regressors, and a renderer based on a soft rasterizer~\cite{liu2019soft}. 

\textbf{Datasets:}
Widely used human body datasets such as CAESAR~\cite{CAESAR} contain high volumes of 3D scans and  body measurements; however these do not come with real images, which must therefore be simulated from the scans with a virtual camera. Recently Yan et al. \cite{yan2020silh} published the \textit{BodyFit} dataset comprising over 4K body scans from which body measurements are computed, and silhouettes are simulated. They also present a small collection of photographs and tape measurements of 194 subjects. To resolve scale, they assume a fixed camera distance. Our BodyM is the first large-scale dataset comprising body measurements paired with silhouettes obtained by applying semantic segmentation on real photographs. To resolve scale, we store height and weight (easy to acquire) rather than assume fixed camera distance (hard to enforce in practice).  

\section{Method} \label{sec:method}

We use the SMPL model~\cite{SMPL:2015} as our basis for adversarial body simulation. SMPL characterizes the human form in terms of a finite number of shape parameters $\beta$ and pose parameters $\theta$. Shape is modeled as a linear weighted combination of basis shapes~(with weights $\beta$) derived from the CAESAR dataset, while pose is modeled as local 3D rotation angles $\theta$ on 24 skeleton joints. SMPL learns a regressor $M(\beta, \theta)$ for generating an articulated body mesh of 6890 vertices from specified shape and pose using \textit{blend shapes}.

\subsection{Body Measurement Estimation Network} \label{subsec:bmnet}

BMnet takes as input either single or multi-view silhouette masks. For single-view, only a frontal segmentation mask is used. For multi-view, the model also leverages the lateral silhouette which provides crucial cues for accurate measurement in the chest and waist areas. Additionally, we use height and weight as input metadata. Height removes the ambiguity in scale when predicting measurements from subjects with variable distance to the camera, while weight provides important cues for body size and shape. Our multi-view measurement estimation network can be written as: 
\begin{equation}
\label{eq:BMnet_mapping}
    y = f_{\psi}(x_f, x_l, \xi, \omega),
\end{equation}
where $x_f$ and $x_l$ are respectively the frontal and lateral silhouettes, $(\xi, \omega)$ are the height and the weight of the subject, and $\psi$ represents network weights.

The network architecture comprises a MNASNet backbone~\cite{tan2019mnasnet} with a depth multiplier of 1 to extract features from the silhouettes. Each silhouette is of size $640\times480$ and the two views are concatenated spatially to form a $640 \times 960$ image. Constant-valued images of the same size representing height and weight are then concatenated depth-wise to the silhouettes to produce an input tensor of dimension $3\times640\times960$ for the network. 
The resulting feature maps from MNASNet are fed into an MLP comprising a hidden layer of 128 neurons and 14 outputs corresponding to body measurements. Unlike previous approaches that attempt the highly ambiguous problem of predicting a high-dimensional body mesh and then subsequently computing the measurements from the mesh~\cite{dibra2016hs}, we directly regress measurements, thus requiring a simpler architecture and obviating the need for storing 3D body mesh ground truth. 

\subsection{Adversarial Body Simulator} \label{subsec:abs}
We present an {\em adversarial body simulator}~(ABS) that searches the latent shape space of the SMPL model in order to find body shapes that are challenging for BMnet. Given a set of shape and pose parameters~($\beta$, $\theta$), we generate a SMPL body mesh~$M(\beta, \theta)$. We then render a 2D silhouette image $x$ of this body using a graphics renderer $R()$, 
given camera parameters $\gamma$ and lighting conditions $\iota$:
\begin{equation}
\label{eq:SMPL_render}
    x = R(M(\beta, \theta), \iota, \gamma).
\end{equation}
Combining Eq.~\ref{eq:BMnet_mapping} and \ref{eq:SMPL_render} we arrive at an expression for measurements predicted by BMnet for a SMPL body as:
\begin{equation}
\label{eq:meas_pred}
    y = f_{\psi}(R(M(\beta, \theta), \iota, \gamma_f), R(M(\beta, \theta), \iota, \gamma_l), \xi, \omega),
\end{equation}
where $y$ is the vector of body measurements predicted by BMnet; $\gamma_f$ are the frontal camera parameters, $\gamma_l$ are the lateral camera parameters where the camera azimuth has been decreased by 90 degrees, and $(\xi, \omega)$ are the height and weight of the subject. The goal of adversarial simulation is to seek challenging inputs that result in high measurement prediction loss $L(y,y_{gt}) = ||y - y_{gt}||^2$ where $y_{gt}$ are ground truth measurements:
\begin{equation}
\label{eq:max_adv}
    \max_{\beta}[L(f_{\psi}(x_f(\beta), x_l(\beta), \xi, \omega),y_{gt})],
\end{equation}
where, $x_f(\beta)$ and $x_l(\beta)$ are the frontal and profile renders with shape parameters $\beta$. We construct our setup so that loss $L$ is differentiable with respect to shapes $\beta$, enabling the use of gradient back-propagation to find adversarial samples. We now investigate in detail the dependence of $y$ and $y_{gt}$ on $\beta$. Turning first to $y$ in Eq.~\ref{eq:meas_pred}, the SMPL model $M$ is linear and thus differentiable with respect to $\beta$. The renderer $R$ is designed as a differentiable projection operator from the 3D body mesh to a 2D silhouette. First, the posed body is lit by a frontal diffuse point light and captured by a perspective camera pointed towards the body mesh. The 2D image is generated using a fully-differentiable soft silhouette rasterizer that aggregates mesh triangle contributions to each 2D pixel in a probabilistic manner ~\cite{liu2019soft}. 
While lighting is not critical for silhouette generation, we include it as part of a general RGB image generation framework.


\begin{figure*}[t]
\centering
\includegraphics[width=0.65\textwidth]{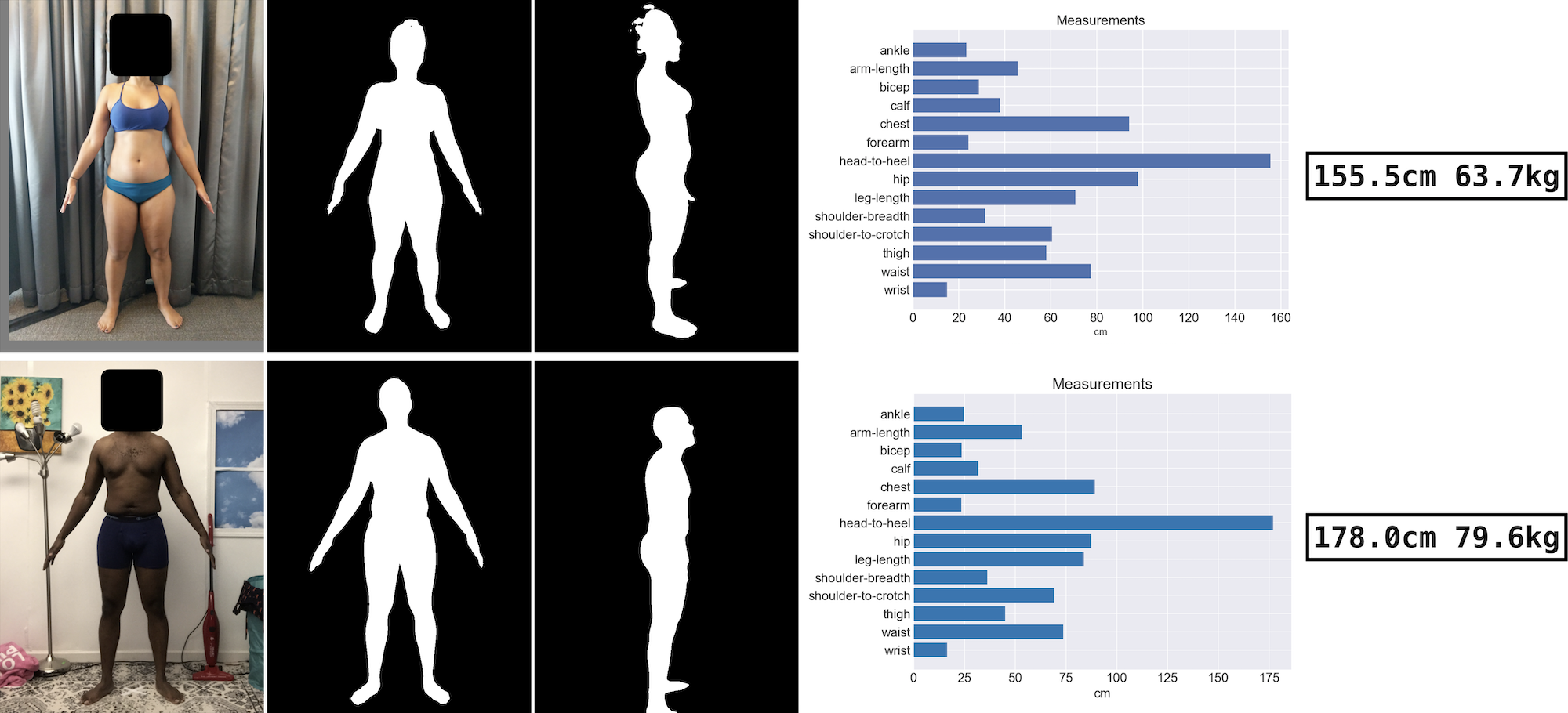}
\caption{Example frontal color photograph, frontal and profile segmentation masks, body measurements and height/weight for different subjects in the BodyM Training Set (top) and the Test-B (bottom) datasets respectively.}
\label{fig:bodym_fig}
\vspace*{-5px}
\end{figure*}

Recall that height and weight ($\xi$, $\omega$) are inputs to BMnet. These are not natural outputs of SMPL; however they are strongly correlated with body shape. We construct a differentiable 3-layer neural network regressor $h$ that predicts height and weight $\xi$ and $\omega$ from shape $\beta$. We train $h$ in a supervised fashion on the CAESAR dataset, which contains subject height and weight as well as body mesh data. We fit a gender-neutral SMPL model with 10 shape parameters ($\beta \in R^{10}$) to the body meshes, providing tuples ($\beta, \xi,  \omega$) for training $h$. The choice of a gender-neutral model is based on our earlier findings that gender contributes minimal improvement to height/weight prediction, and the fact that gender must be determined either automatically (which is error prone) or by asking the user (not everyone shares or identifies with gender).  Average prediction errors of $h$ on independent test sets are within 1 cm and 1 kg respectively.   

Next we turn to $y_{gt}$. For a given SMPL body mesh $M$ the 14 body measurements are obtained by computing the lengths of curves traversing pre-specified vertex paths on the mesh. These curve lengths are computed by summing vertex-to-vertex distances along the path. This operation, denoted $y_{gt}(\beta) = g(M(\beta, \theta))$, is the same used to annotate the BodyM dataset (see Sec. \ref{sec:bodym}). The fully differentiable silhouette renderer is shown in Figure~\ref{fig:renderer}.

In order to sample adversarial bodies we 
optimize $\beta$ by gradient ascent, backpropagating the gradient of the loss with respect to $\beta$:
\begin{equation}
    \nabla_\beta L[f_{\psi}(x_f(\beta), x_l(\beta), \xi(\beta), \omega(\beta)), y_\text{gt}(\beta)],
\end{equation}
where 
height and weight depend on $\beta$ via $h()$, and $y_{gt}$ depends on $\beta$ via $g()$. For body shape analysis (Sec. \ref{subsec:abs_analysis}) we fix pose $\theta$  to a canonical A-pose, while for  training BMnet, we sample $\theta$ randomly from poses of real humans in the BodyM dataset. Henceforth we omit pose, camera and lighting parameters for brevity. The $\beta$ are updated using the gradient ascent update rule:
\begin{equation}
    \beta_{k+1} = \beta_{k} + \eta \nabla_\beta L[f_{\psi}(x_f(\beta), x_l(\beta), \xi(\beta), \omega(\beta)), y_\text{gt}(\beta)],
    \label{eq:beta_update}
\end{equation}
where $\eta$ is a weight hyperparameter. Note that only $\beta$ is updated, and the weights $\psi$ of the model $f$ are fixed. 
We illustrate the optimization in Figure~\ref{fig:adv_shape_opt}.

\textbf{Adversarial augmentation:} 
In order to train a network in an adversarial manner, we need to ensure our adversarial sampling selects diverse yet realistic body shapes. For this purpose, when doing adversarial augmentation, we initialize $\beta$ by selecting at random shape parameters that have been fitted to real human bodies in the BodyM training set. We then optimize these shapes for $k$ iterations following the update rule shown in Equation~\ref{eq:beta_update}. This yields samples that are challenging, yet close to real body shapes. An alternate strategy would be to sample only around challenging examples in the training set; however we have found that excessive emphasis on hard examples causes BMnet to overfit on these and compromise mean performance.
%
%


We first pre-train BMnet on real examples from BodyM, and then fine-tune for 10 epochs using synthetic examples from the aforementioned augmentation. 
Training BMnet minimizes the L1 difference between regressed and target measurements.
Synthetic bodies are not repeated over epochs, so that in 10 epochs the network sees roughly 10 times more data than the one using real data. Finally we perform another fine-tuning on the real BodyM data to bridge the synthetic-to-real domain gap. We note that synthetic silhouettes produced by the renderer $R$  are noise-free, while silhouettes in BodyM are generated by segmenting real RGB photos, and thus contain realistic noise artifacts.  
Our augmentation strategy is inspired by adversarial training using pixel-level adversarial attacks~\cite{madry2018towards}, with some key differences: (1) we search through interpretable parameters of a simulator to find adversarial samples instead of modifying image pixels using high-frequency noise; (2) we use a gradient descent update instead of the quantized fast gradient sign update rule, since the latter leads to a coarse exploration of the landscape that is not suitable when searching for simulated adversarial examples in shape and pose space. Note that some additional training computational cost exists, but is in the order of 1\%.

\section{BodyM Dataset} \label{sec:bodym}
Synthetic datasets used in previous body measurement work often lack the detail and diversity of real body shapes. To address this domain gap, we introduce \textit{BodyM}, the first public dataset containing 8,978 frontal and lateral silhouette photos paired with height, weight and 14 body measurements for 2,505 real individuals. 
The ethnicity distribution of BodyM is: White 40\%, Asian 30\%, Black/African American 14\%, American Indian or Alaska Native 1\%, Other 15\%; with 15\% of the individuals also indicating Hispanic.
The training-test breakdown is reported in 
Table~\ref{table:bodymdataset} (top).
Table~\ref{table:bodymdataset} (bottom) reports gender and BMI statistics. We note that \textit{BMI $\in$ 18.5-25} and \textit{BMI $\in$ 25-30} are the dominant body shape categories. RGB photos were captured in a well-lit, indoor setup, with subjects standing in A-Pose wearing tight-fitting clothing, as shown in Figure~\ref{fig:bodym_fig}. Capture distance varied between 5.5-6.5 feet. Silhouettes were obtained by applying semantic segmentation on RGB~\cite{Chen_2018_ECCV}, thus exhibiting realistic segmentation artifacts not found in existing simulated datasets (e.g., Figure~\ref{fig:bodym_fig} top)
3D scans of each subject were acquired with a Treedy photogrammetric scanner, registered to the SMPL mesh topology, and reposed to a canonical ``A-pose". The following body measurements were then computed on the meshes using the procedure described in Sec.~\ref{subsec:abs}: \textit{ankle girth, arm-length, bicep girth, calf girth, chest girth, forearm girth, head-to-heel length, hip girth, leg-length, shoulder-breadth, shoulder-to-crotch length, thigh girth, waist girth,} and \textit{wrist girth}.

\begin{table}[th]
\centering
\begin{subtable}[t]{0.6\columnwidth}
\centering
\resizebox{\columnwidth}{!}{
\begin{tabular}{lclclcl}
\toprule
                  & \multicolumn{2}{c}{\textbf{Train}} & \multicolumn{2}{c}{\textbf{Test-A}} & \multicolumn{2}{c}{\textbf{Test-B}} \\ 
\midrule
\textbf{Subjects} & \multicolumn{2}{c}{2,018}           & \multicolumn{2}{c}{87}           & \multicolumn{2}{c}{400}           \\
\textbf{Silhouettes}      & \multicolumn{2}{c}{6,134}           & \multicolumn{2}{c}{1,684}          & \multicolumn{2}{c}{1,160}  \\
\bottomrule
\end{tabular}
}
\end{subtable}  
\begin{subtable}[t]{\columnwidth}
\centering
\resizebox{0.8\columnwidth}{!}{
\begin{tabular}{@{}lrrrrrrr@{}}
\toprule
\multicolumn{1}{c}{}                   & \multicolumn{2}{c}{\textbf{Training Set}}         & \multicolumn{2}{c}{\textbf{Test-A Set}}       & \multicolumn{2}{c}{\textbf{Test-B Set}}              \\ 
\midrule
\multicolumn{1}{c}{\textbf{}}          & \textbf{Male}           & \textbf{Female}         & \textbf{Male}           & \textbf{Female}         & \textbf{Male}           & \textbf{Female}          \\ 
\midrule
\textbf{GENDER}                        & 60\%                    & 40\%                    & 52\%                    & 47\%                    & 39\%                    & 61\%                     \\
\textbf{BMI \textless{}18.5}           & 0\%                     & 2\%                     & 1\%                     & 3\%                     & 1\%                     & 4\%                      \\
\textbf{BMI 18.5-25}                   & 28\%                    & 23\%                    & 33\%                    & 31\%                    & 18\%                    & 37\%                     \\
\textbf{BMI 25-30}                     & 25\%                    & 9\%                     & 15\%                    & 7\%                     & 13\%                    & 10\%                     \\
\textbf{BMI 30-40}                     & 6\%                     & 5\%                     & 2\%                     & 6\%                     & 7\%                     & 7\%                      \\
\textbf{BMI 40-50}                     & 0\%                     & 1\%                     & 0\%                     & 0\%                     & 1\%                     & 3\%                      \\
\textbf{BMI \textgreater{}=50}         & 0\%                     & 0\%                     & 0\%                     & 0\%                     & 0\%                     & 0\%                      \\
\bottomrule
\end{tabular}}
\end{subtable}  
\caption{BodyM dataset statistics (top), gender and BMI statistics for the BodyM (bottom).
\label{table:bodymdataset}}
\vspace{-10pt}
\end{table}

For the training and Test-A sets, subjects were photographed and 3D-scanned by lab technicians. For the Test-B set, subjects were scanned in the lab, but photographed in a less-controlled environment with diverse camera orientations and lighting conditions, to simulate in-the-wild image capture. For privacy reasons, we do not release the original RGB images (not anyway needed by BMnet). 

\section{Experimental Results}

For all experiments, unless noted, we train the baseline BMnet for 150k iterations on the BodyM training set using the Adam optimizer with a learning rate of $10^{-3}$ and a batch size of 22. We select the best model using a validation set corresponding to 10\% of the training data. The learning rate follows a multi-step schedule, whereby we reduce the learning rate at $75\%$ and $88\%$ of the training.

\textbf{Metrics:} We define measurement accuracy based on quantiles of absolute measurement errors. TP90 (TP75, TP50) metrics are defined by computing the 90th (75th, 50th) quantile cutoff for all 14 measurements, and reporting the mean of these values. Mean absolute error (MAE) is reported for selected experiments.

\subsection{Adversarial Body Shape Analysis}
\label{subsec:abs_analysis}
We use ABS to reveal regimes of the body shape space where a pre-trained BMnet performs poorly. We initialize a 10-dimensional SMPL shape vector $\beta$ to fall randomly within a small ball of radius $0.01$ around the zero-vector. We then iteratively update $\beta$ to maximize BMnet loss. 
The camera parameters $\gamma$ are chosen to mimic the setup used to capture real images in BodyM. The lighting parameters $\iota$ represent a point illumination source that shines directly onto the subject from behind the camera, using only diffuse lighting, in order to avoid specular artifacts from corrupting the silhouette. While we fix pose, lighting and camera parameters as constants in this experiment, we note that our framework can be readily generalized to adversarial sampling of all these parameters.

For ABS we use adversarial sampling with a learning rate $\eta$ of $0.1$ and $k=10$. Shape parameters are clamped in a $[-3,3]$ range to prevent unrealistic body shapes. We compare samples generated using ABS to random body samples. Using this random sampling, we sample the shape space uniformly in the $[-3,3]$ range. Our rationale is that in the absence of prior knowledge about $f$, the uniform distribution is the maximum entropy distribution, hence providing the strongest sampling baseline. 

\begin{figure*}[t]
  \centering
    \begin{subfigure}{0.8\columnwidth}
    \includegraphics[width=\columnwidth]{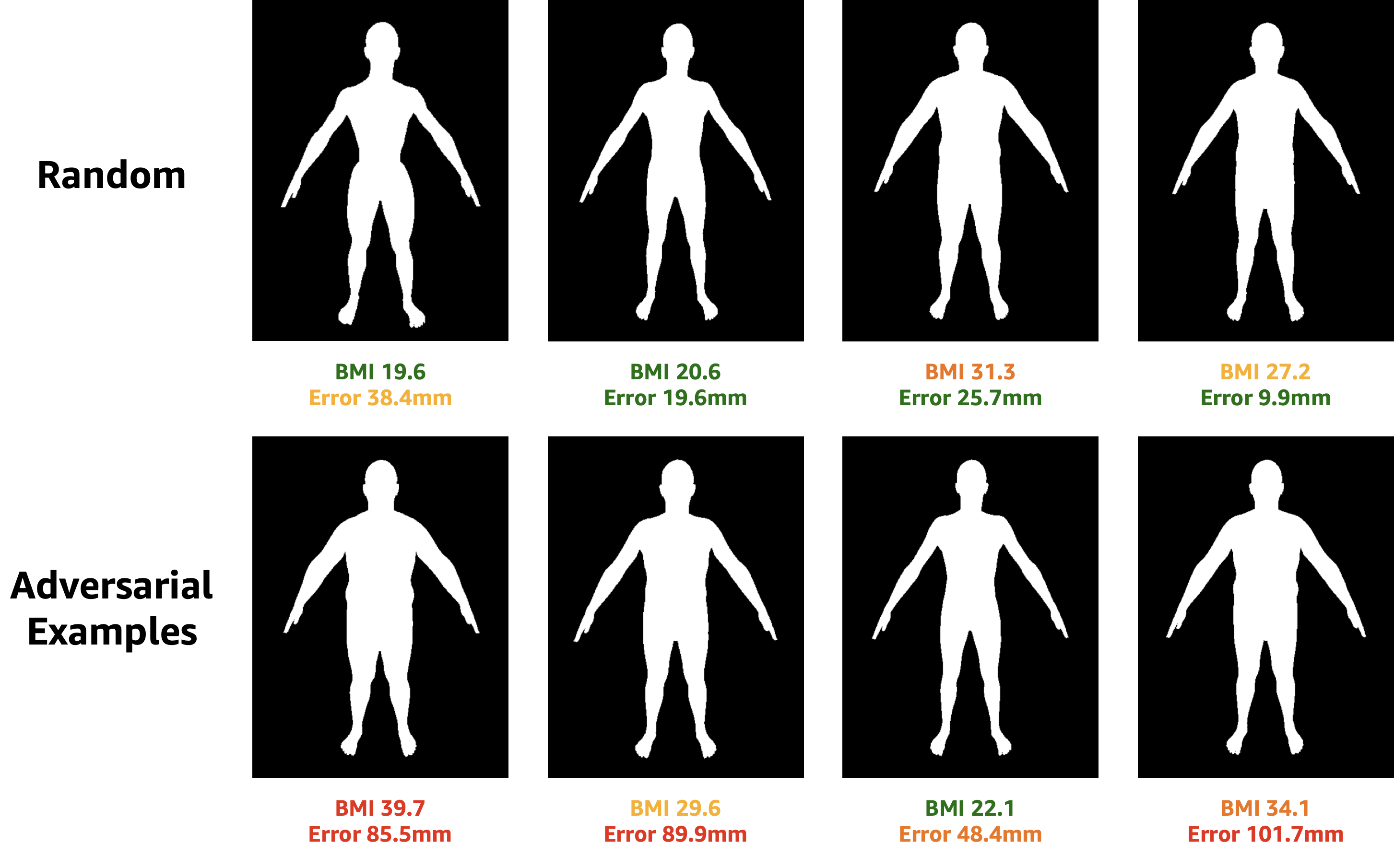}
  \end{subfigure}
  \hspace{10pt}
  \begin{subfigure}{0.8\columnwidth}
    \includegraphics[width=\columnwidth]{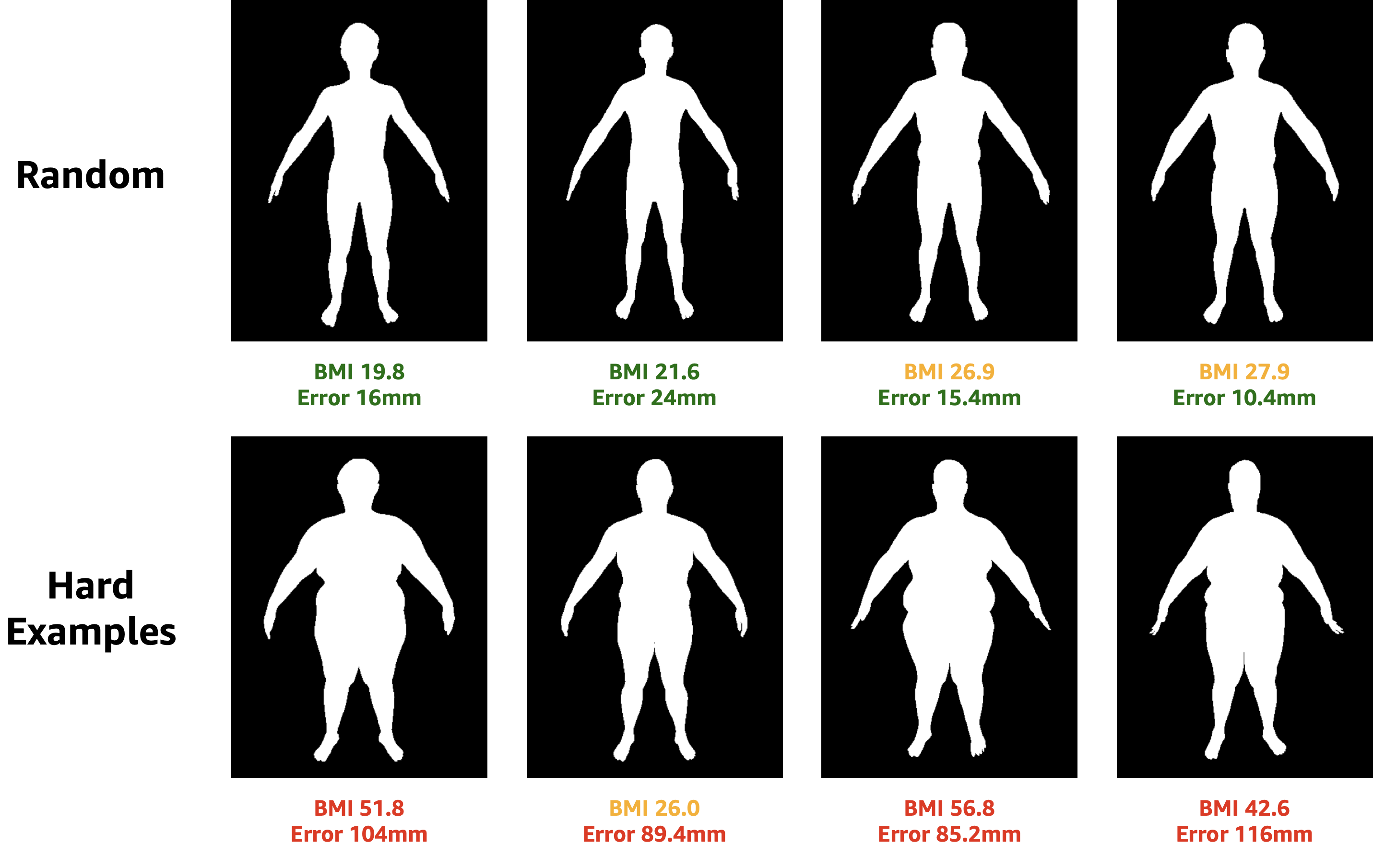}
  \end{subfigure}
  \caption{Comparison of randomly vs. adversarially simulated bodies (left). Comparison of random sampling vs. hard examples of real bodies with high body measurement estimation error (right).
  \label{fig:adv_samples}}
  \vspace*{-3pt}
\end{figure*}

We show qualitative comparisons between randomly simulated bodies and adversarially simulated bodies in Figure~\ref{fig:adv_samples}, left. 
We observe that adversarial body shapes are of high BMI compared to random bodies. The mean measurement error of $f$ for the adversarial bodies is also much higher than that for random bodies. We also note adversarial samples that are not of high BMI but with high measurement error~(third sample in Fig.~\ref{fig:adv_samples}, left). 
Fig.~\ref{fig:adv_samples}, right, shows examples of real bodies, both random and samples with high error. We observe that the error for the hard samples is similar to that of the simulated adversarial samples. Furthermore, we can see that challenging samples in the real world are also of high BMI, similar to our simulated adversarial bodies. Aggregating this analysis over the entire population, the average BMI's for the random and adversarial body groups are 28.1 and 35.8 respectively; and the mean measurement errors in millimeters~(mm) for the two groups are 34.8 and 92.2. 

\begin{figure}[t]
\centering
\includegraphics[width=0.8\columnwidth]{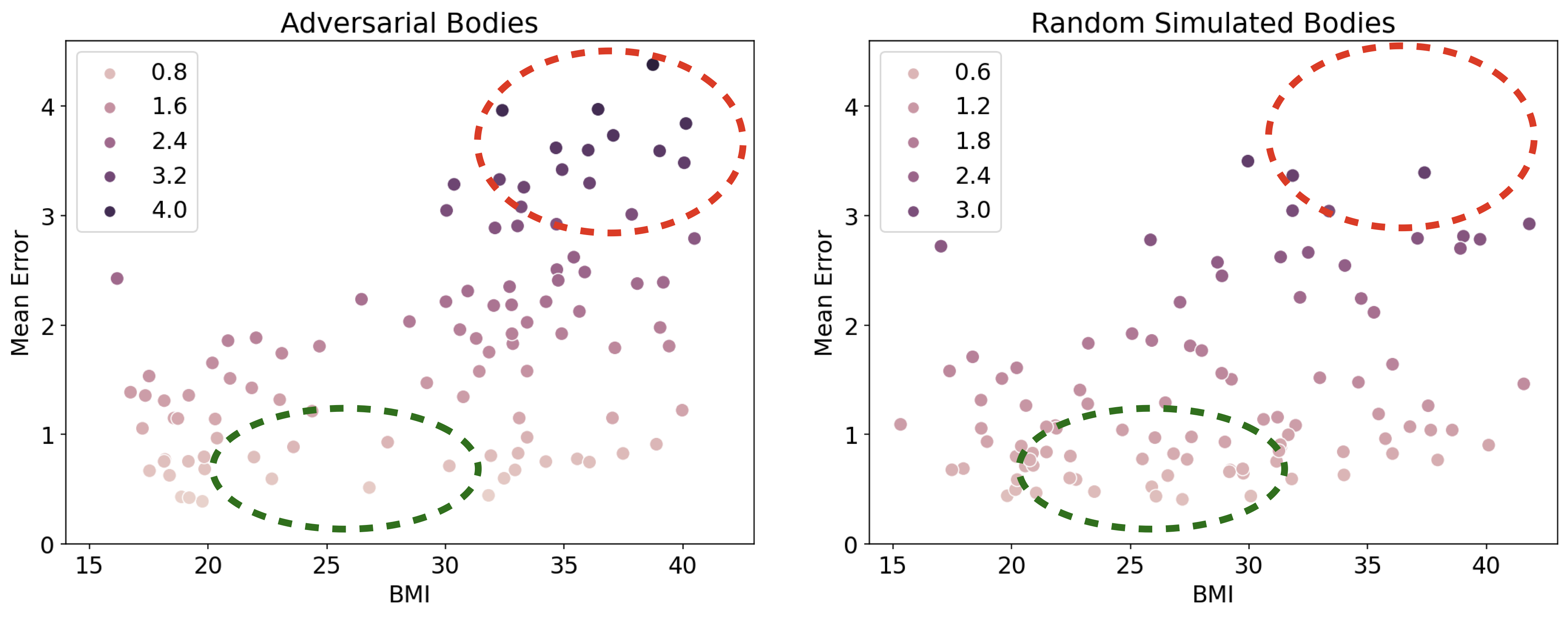}
\caption{Prediction error vs. BMI for adversarial (left) and random (right) body sampling. The adversarial scheme selects more high-error and high-BMI samples (red ellipse) than the random sampling, more concentrated in low-error low-BMI areas (green ellipse).
\label{fig:bmi_analysis}}
\vspace*{-5pt}
\end{figure}

\begin{figure}[t]
\centering
\includegraphics[width=0.8\columnwidth]{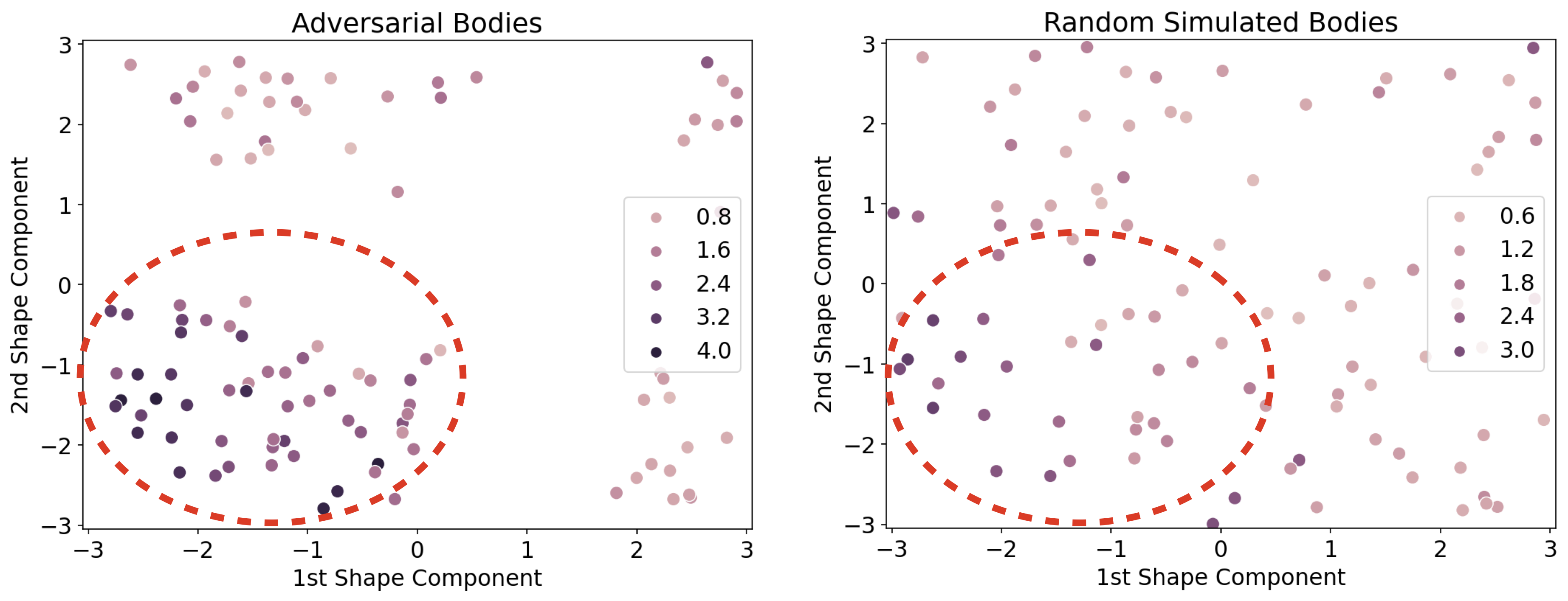}
\caption{Distribution of adversarial (left) vs. random (right) sampling along the first two components of SMPL shape space. Adversarial bodies are more concentrated in the negative quadrant (red ellipse).
\label{fig:shape_analysis}
}
\vspace*{-15pt}
\end{figure}

\begin{figure}[t]
\centering
\includegraphics[width=0.6\columnwidth]{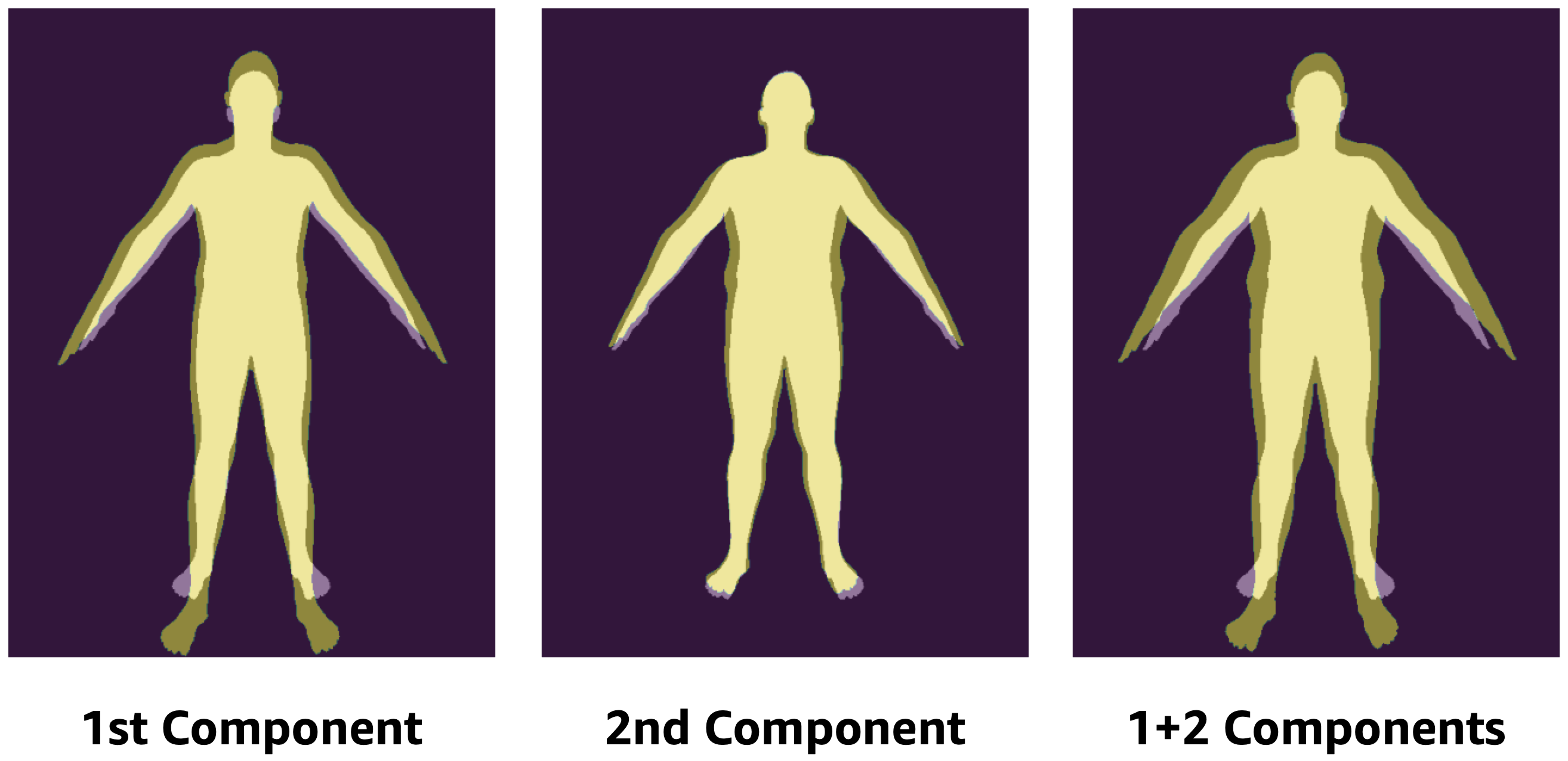}
\caption{Changes in body shape (from yellow to green) from adding small negative perturbations to the 1st and 2nd shape components. We surmise that the network underperforms with taller, larger, and wider bodies.
\label{fig:shape_analysis_bodies}}
\vspace*{-15pt}
\end{figure}

As another visualization, Figure~\ref{fig:bmi_analysis} plots mean measurement error vs. BMI for adversarially sampled and random bodies. Error magnitudes are color-coded. We observe that the adversarially sampled population contains more bodies with higher error~(red circles) and fewer bodies with low error~(green circles). Furthermore, the adversarially sampled population contains many more samples with high BMI, which seem to directly contribute to higher mean error. In Figure~\ref{fig:shape_analysis} we visualize adversarial and random body sampling in the first two principal dimensions of the latent SMPL shape space. Again error magnitudes are color-coded. Adversarial~(and high-error) bodies are largely concentrated in the negative quadrant of the shape space. For visual interpretation, Figure~\ref{fig:shape_analysis_bodies} shows that negative perturbations in $\beta_1$ and $\beta_2$  result in taller and wide bodies.

\subsection{Ablation Studies}
We highlight the impact of key elements of our body measurement estimation architecture in Table~\ref{table:ablation}. First we study the effects of using one~(frontal) input silhouette vs. two~(frontal and lateral), as well as the effect of adding height and weight as metadata inputs to the network. 

\begin{table}[t]

\centering
\resizebox{\columnwidth}{!}{%
\begin{tabular}{lccccccc}
\toprule
& \multicolumn{3}{c}{\bf Overall} & \multicolumn{1}{c}{\bf Chest} & \multicolumn{1}{c}{\bf Hip} & \multicolumn{1}{c}{\bf Waist} \\
\midrule
{\bf } & TP90 & TP75 & TP50 & MAE & MAE & MAE \\ 
\midrule
Single-View & 41.91 & 29.13 & 17.09 & 33.95 & 31.03 & 31.93 \\
Multi-View & 39.02 & 26.55 & 14.85 & 28.66 & 28.29 & 27.32 \\
Multi-View + Height & 20.21 & 13.87 & 8.00 & 19.38 & 15.97 & 18.71 \\
Multi-View + Weight & 18.55 & 12.62 & \textbf{7.20} & \textbf{15.22} & 10.54 & \textbf{13.69} \\
Multi-View + Height + Weight & \textbf{18.42} & \textbf{12.55} & 7.34 & 15.92 & \textbf{9.74}& 15.44 \\
\bottomrule
\end{tabular}
}
\caption{Ablations on single- vs. multi-view and height/weight inputs (errors in mm.) on BodyM TestA. Addition  of  a  second view  improves  the accuracy of body measurements.  Adding  only  the  weight  has  stronger (positive) impact than adding only height. Robustness to outliers is improved when adding both height and weight.
\label{table:ablation}}
\vspace{-10pt}
\end{table}

\begin{table}[t]
\resizebox{\columnwidth}{!}{%
\begin{tabular}{lcccccc}
\toprule
& \multicolumn{3}{c}{\bf Overall} & \multicolumn{1}{c}{\bf Chest} & \multicolumn{1}{c}{\bf Hip} & \multicolumn{1}{c}{\bf Waist} \\
\midrule
{\bf }  & TP90 & TP75 & TP50 & MAE & MAE & MAE \\ 
\midrule
Single-View (No Aug.) & 19.10 & 13.00 & 7.64 & 19.18 & 11.53 & 16.12 \\
Single-View (Random Aug.) & 18.98 & 12.84 & 7.50 & 19.13 & 11.43 & \textbf{15.76} \\
Single-View (Adv. Aug.) & \textbf{18.90} & \textbf{12.82} & \textbf{7.44} & \textbf{18.84} & \textbf{11.14} & 15.78 \\
\midrule
Multi-View (No Aug.) & 16.45 & 11.06 & 6.51 & \textbf{14.40} & 10.88 & 13.40 \\
Multi-View (Random Aug.) & 16.43 & 11.06 & \textbf{6.48} & 14.66 & 10.60 & 13.17 \\
Multi-View (Adv. Aug.) & \textbf{16.00} & \textbf{10.00} & 6.53 & 14.52 & \textbf{10.00} & \textbf{13.00}\\
\midrule
\midrule
Multi-View (No Aug.) & 26.52 & 17.64 & 10.04 & 24.60 & 19.55 & 21.75 \\
Multi-View (Random Aug.) & 26.13 & 17.35 & 9.90 & 23.09 & 18.87 & 22.44 \\
Multi-View (Adv. Aug.) & \textbf{25.00}& \textbf{16.28} & \textbf{9.50} & \textbf{22.98} & \textbf{18.09} & \textbf{21.10} \\
\bottomrule
\end{tabular}
}
\caption{Ablations (on BodyM TestA) for synthetic data augmentation strategies. BMNet trained on the full training set (top two row blocks) and a reduced training set (bottom row block). Adversarial augmentation achieves lower errors (up to {\em 10\%}) over no augmentation or random sampling.
\label{table:advsampling_eval_bodym}
}
\centering
\vspace{-10pt}
\end{table}

\begin{table*}[t]
\centering
\resizebox{0.9\textwidth}{!}{%
\begin{tabular}{lcccccccccccccc|c}
\toprule
{\bf } & Ankle & Arm & Bicep & Calf & Chest & Forearm & H2H & Hip & Leg & S-B & S-to-C & Thigh & Waist & Wrist & Overall \\ 
\midrule
Ours No Aug. & 7.89 & 9.97 & 11.42 & 11.29 & 24.60 & 7.31 & 10.10 & 19.55 & 14.33 & 7.83 & 9.72 & 15.54 & 21.75 & 5.73 & 12.65 \\
Ours Adv. Aug. &\textbf{ 7.59} & \textbf{9.91} & \textbf{11.26} & \textbf{10.88} & \textbf{22.98} & \textbf{7.16} & \textbf{9.19} & \textbf{18.09} & 14.97 & \textbf{7.67} & \textbf{9.30} & \textbf{14.11} & \textbf{21.10} & \textbf{5.52} & \textbf{12.12} \\
\bottomrule
\end{tabular}
}
\caption{Mean average individual measurement errors on Test-A using the reduced training set (bottom). \it{H2H} stands for Head-to-Heel, \it{S-B} is Shoulder-Breadth and S-to-C is \it{Shoulder-to-Crotch}.
\label{table:individual_meas_aug}
}
\centering
\vspace{-10pt}
\end{table*}

\begin{table*}[t]
    \centering
    \resizebox{0.8\textwidth}{!}{%
\begin{tabular}{@{}lcccccccccccccc|c@{}}
\toprule
     & Ankle & Arm-L  & Bicep        & Calf         & Chest        & Forearm      & H2H & Hip          & Leg-L   & S-B   & S-to-C & Thigh        & Waist        & Wrist        & Overall    \\ \midrule
Dibra et al.~\cite{dibra2016hs} & 2.0   & 2.7          & 3.3          & 3.3          & 7.2          & 2.3          & 4.0          & {\em 6.0}          & 2.8          & 2.9          & 2.9           & 4.9          & 8.1          & 2.0          & 3.78          \\
Smith et al.~\cite{Smith2019TowardsA3} & 2.1   & \textbf{1.7} & 2.7          & 2.3          & 4.7          & 1.9          & \textbf{2.3} & 3.0          & \textbf{1.5} & 1.9          & \textbf{1.5}  & 2.4          & 4.8          & 2.5          & 2.72          \\
Ours & \textbf{0.8}   & 1.9          & \textbf{1.7} & \textbf{0.8} & \textbf{4.6} & \textbf{1.3} & 3.6          & \textbf{\em 1.8} & 2.1          & \textbf{0.9} & 1.9           & \textbf{1.7} & \textbf{3.8} & \textbf{0.7} & \textbf{1.97} \\ 
\bottomrule
\end{tabular}
}
\caption{MAE (mm) for different methods on the simulated dataset from \cite{Smith2019TowardsA3}. {\it H2H} stands for Head-to-Heel, {\it Arm-L} and {\it Leg-L} is arm/leg length, {\it S-B} is Shoulder-Breadth and S-to-C is {\it Shoulder-to-Crotch}.  Ours achieves up to {\em 70\%} error reduction on real-body measurements.
\label{table:benchmarking_simulated}}
\vspace{-10pt}
\end{table*}

\begin{table*}[t]
\centering
\resizebox{0.8\textwidth}{!}{%
\begin{tabular}{lcccccccccccccc|c}
\toprule
{\bf } & Neck & Chest & Waist & Pelvis & Wrist & Bicep & Forearm & Arm & Leg & Thigh & Calf & Ankle & Height & Shoulder & Overall \\ 
\midrule
Yan et al.~\cite{yan2020silh} & 11.8 & 23.0 & 16.5 & \textbf{13.3} & 4.1 & 11.4 & 7.2 & \textbf{7.6} & \textbf{9.2} & 17.8 & 8.8 & \textbf{5.4} & 9.0 & 9.2 & 11.0 \\
Ours & \textbf{11.0} & \textbf{15.2} & \textbf{15.7} & 17.3 & \textbf{3.8} & \textbf{4.7} & \textbf{3.9} & 7.7 & 10.0 & \textbf{7.5} & \textbf{8.6} & 10.0 & \textbf{13.4} & \textbf{7.1} & \textbf{9.7} \\
\midrule
Yan et al.~\cite{yan2020silh} & {\em 14.6} & 21.7 & 17.1 & 14.7 & 5.2 & 9.3 & 8.5 & \textbf{6.4} & \textbf{6.5} & \textbf{11.6} & \textbf{9.2} & \textbf{6.1} & \textbf{8.6} & 7.6 & 10.5 \\
Ours & \textbf{4.4} & \textbf{9.1} & \textbf{10.8} & \textbf{7.7} & \textbf{5.2} & \textbf{3.9} & \textbf{5.3} & \textbf{6.4} & 10.2 & 13.2 & 9.8 & 12.2 & 20.7 & \textbf{6.5} & \textbf{9.0} \\
\bottomrule
\end{tabular}
}
\caption{MAE (mm) on Body-Fit~\cite{yan2020silh} for male (top) and female (bot.) bodies with error reduction up to {\em 70\%} using ours.
\label{table:benchmarking_bodyfit}
}
\vspace{-10pt}
\end{table*}

\begin{table*}[t!]
\centering
\resizebox{0.6\textwidth}{!}{%
\begin{tabular}{lccccccc}
\toprule
& \multicolumn{3}{c}{\bf Overall} & \multicolumn{1}{c}{\bf Chest} & \multicolumn{1}{c}{\bf Hip} & \multicolumn{1}{c}{\bf Leg Length} & \multicolumn{1}{c}{\bf Waist} \\
\midrule
{\bf } & TP90 & TP75 & TP50 & MAE & MAE & MAE & MAE \\ 
\midrule
SPIN~\cite{kolotouros2019learning} & 81.10 & 57.33 & 33.96 & 74.45 & 65.41 & 35.81 & 77.39 \\
STRAPS~\cite{sengupta2020synthetic} & 103.61 & 75.74 & 45.67 & 82.30 & 63.96 & 48.71 & 108.00 \\
Sengupta et al.~\cite{Sengupta_2021_ICCV} & 68.81 & 47.64 & 28.71 & 53.07 & 47.43 & 42.11 & 53.20 \\
Ours (Single-View, No Metadata) & \textbf{41.91} & \textbf{29.13} & \textbf{17.09} & \textbf{33.95} & \textbf{31.03} & \textbf{25.80} & \textbf{31.93} \\
\bottomrule
\end{tabular}
}
\caption{Performance comparison for measurement estimation using different methods on the BodyM dataset. 
\label{table:benchmarking_bodym}
}
\vspace{-10pt}
\end{table*}

The addition of a second~(lateral) view improves results by providing additional evidence not found in the frontal view. The addition of height and weight dramatically affect the network's ability to correctly predict measurements. Adding only the weight has stronger impact than adding only height. Robustness to outliers is improved when adding both height and weight, evidenced by the TP90 and TP75 metrics, although some specific measurements are less accurate than when only using one input. This could be attributed to slight noise in height and weight.

Next we evaluate the impact of augmenting real data samples with synthetic data drawn from different sampling strategies when training BMnet. We fit the SMPL model with 10 shape and 72 pose parameters to each real body in the BodyM training set. We then sample around these real parameters, in order to create synthetic augmentations with shape and pose in the vicinity of real bodies. We compare two methods of augmentation: random sampling and the proposed adversarial sampling around the BodyM shape parameters. Random sampling is performed uniformly within a hypercube with side length of 0.5 around the real shape parameters. For adversarial sampling, we initialize shape using the BodyM parameters and optimize using the Adam optimizer for 5 steps using a learning rate of 0.1.

We also evaluate a baseline network trained only on real BodyM data. All evaluations are performed on the independent BodyM Test-A set, and results are reported in Table~\ref{table:advsampling_eval_bodym}. We observe that for the single-view scenario, random augmentation achieves slightly better results than no augmentation, and adversarial augmentation further improves results over random augmentation, with no additional data acquisition cost. This trend is consistent across aggregate TP metrics, most of the individual body measurements, and the multi-view scenario. We see relatively uniform improvements over different BMI categories.

We repeat the ablation study with a reduced training dataset of 1K randomly chosen real samples ($\sim$ 10\% percent of the data used in the previous experiment). Training of BMnet thus weighs more heavily on synthetic augmentation. Results are shown in the lowest block of Table~\ref{table:advsampling_eval_bodym}. Adversarial augmentation produces the best overall performance, with significant improvements on several individual measurements. This highlights the benefit of adversarial sampling in scenarios where real data is limited. Finally, in Table~\ref{table:individual_meas_aug}, we break down the performance of multi-view BMnet by individual measurements, with and without adversarial augmentation, in the reduced-data scenario. Adversarial augmentation produces a noticeable decrease in almost every individual body measurement error. Results for the Test-B set are included in Sup. Mat.

\subsection{State-of-the-Art Comparisons}
We compare our method with recent state-of-the-art body measurement approaches by Smith et al.~\cite{Smith2019TowardsA3} and Dibra et al.~\cite{dibra2016hs} on the simulated test set taken from \cite{Smith2019TowardsA3}. Both these techniques tackle body measurement estimation under settings similar to ours, where inputs are silhouettes, pose is constrained, and  body shape is highly variable. One difference is that our method and \cite{Smith2019TowardsA3} directly incorporate height and weight inputs, while \cite{dibra2016hs} infer height by assuming subjects are captured at a fixed distance. 
In Table~\ref{table:benchmarking_simulated}, we evaluate our method with adversarial augmentation by ABS, and follow the testing protocols described in \cite{Smith2019TowardsA3}, comparing our method's performance directly with numbers reported in the respective references.
Our method outperforms both alternatives in terms of overall mean error and 10 out of 14 individual measurements, often by a significant margin~(up to 91\% in the reduction of mean errors). 

Table~\ref{table:benchmarking_bodyfit} compares our method with Yan et al. \cite{yan2020silh} on real human bodies in their \textit{BodyFit} dataset. We train multi-view BMnet on \textit{BodyFit} and adapt it to exclude height and weight inputs.  Errors for their approach are taken directly from their paper. We outperform \cite{yan2020silh} on the majority of measurements, as well as the overall average error, demonstrating robustness of BMnet across different datasets.

Finally, for completeness, we compare our method with recent 
human mesh reconstruction (HMR) methods, SPIN~\cite{kolotouros2019learning}, STRAPS~\cite{sengupta2020synthetic}, and Sengupta et al.~\cite{Sengupta_2021_ICCV} on BodyM Test-A. We compute body measurements from HMR by using the measuring function $g$ on the predicted mesh. For fair comparison, we evaluate our technique with a single input and without height and weight metadata. 
Table~\ref{table:benchmarking_bodym} shows that our method reduces most errors by more than 35\%.
This error reduction is due to the fact that our network directly regresses measurements and that we have measurement supervision from the large training corpus in BodyM - we compare against methods with direct measurement prediction in Tables \ref{table:benchmarking_simulated} and \ref{table:benchmarking_bodyfit}. Note in comparing Tables \ref{table:benchmarking_simulated} and \ref{table:benchmarking_bodym} that errors on real human data in BodyM are substantially higher than those on simulated data; demonstrating that BodyM provides a new challenging benchmark.

\section{Conclusions}
We propose BMnet to estimate body measurements from silhouettes, height and weight. The key contribution is a fully differentiable adversarial training scheme generating challenging bodies within the SMPL shape space, and revealing potential training gaps. When BMnet training is augmented with these adversarial shapes, measurement accuracy improves on real humans, producing new state-of-the-art results, particularly when real data is limited. We release BodyM, a new challenging large-scale dataset acquired with real human subjects to promote progress in body measurement research.




{\small
\balance
\bibliographystyle{ieee_fullname}
\bibliography{egbib}
}

\newpage
\nobalance
\input{supp}

\end{document}

%% file: supp.tex
\appendix
\section{Appendix}

\subsection{Body Shape Analysis}
\begin{figure}[h]
\centering
\includegraphics[width=\columnwidth]{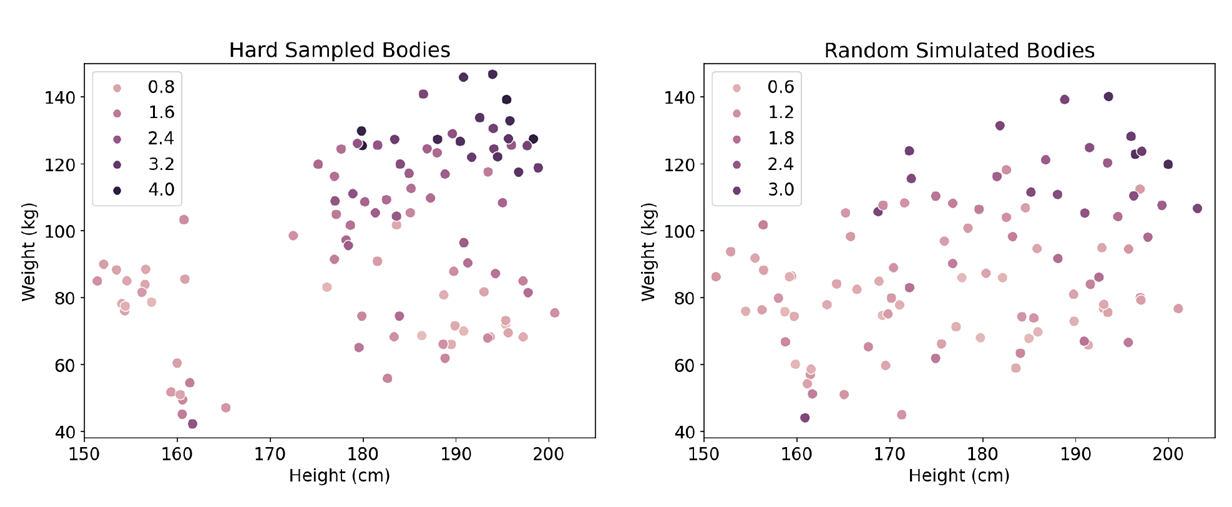}
\vspace*{-1.5em}
\caption{Distribution of adversarial (left) vs. random (right) sampling in terms of weight vs. height. Adversarial bodies are more concentrated in regions of higher weight and height.
}
\label{fig:weight_vs_height}
\vspace*{-1em}
\end{figure}

\begin{figure}[h]
\centering
\includegraphics[width=\columnwidth]{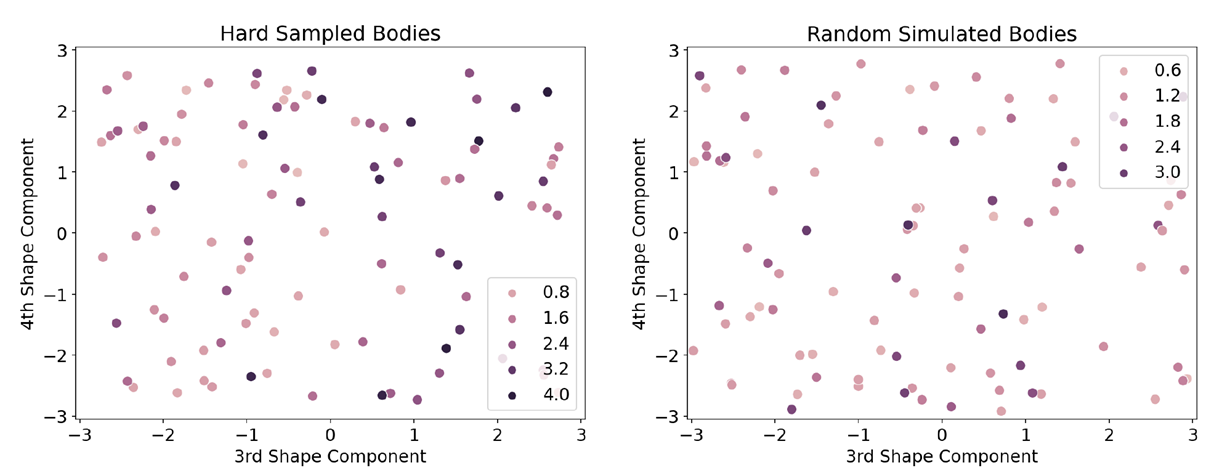}
\vspace*{-1.5em}
\caption{Distribution of adversarial (left) vs. random (right) sampling in terms of 4th vs 3rd principal component ($\beta$) of SMPL shape. Adversarial and random sampling are not clearly distinct along these two dimensions.
}
\label{fig:beta4_vs_beta3}
\vspace*{-1em}
\end{figure}

We include additional analysis of body shapes generated by ABS vs. random sampling. Fig.~\ref{fig:weight_vs_height} plots weight vs. height for adversarial and random sampling. Consistent with analysis in Sec. 5.1 of the paper, we note that adversarial bodies are more densely concentrated in regions of greater weight and height. Fig.~\ref{fig:beta4_vs_beta3} plots the 3rd vs 4th SMPL~\cite{SMPL:2015} shape components. Unlike the first two components, these two variables do not distinctly separate ABS from random sampling, indicating that rare body shapes manifest themselves more strongly along some shape components than others.

\subsection{Additional Experimental Evaluation}

We report additional results comparing our adversarial augmentation with an augmentation-free baseline for training BMnet. In Table~\ref{table:supp_testb_minimal} we show results on the minimal-clothing TestB subset. The findings are consistent: adversarial augmentation improves results in terms of both overall and individual metrics. In Table~\ref{table:supp_measurements_normal} we show mean errors for the individual  body measurements from TestA. Adversarial sampling improves accuracy for 11 out of 14 measurements. The same analysis is performed on the reduced training scenario on TestA  ($\sim$ 10\% of the full training set) in the main manuscript. The gains from adversarial augmentation are even stronger in this scenario.

\begin{table*}[t]
\centering
\resizebox{0.7\textwidth}{!}{%
\begin{tabular}{lccccccccc}
\toprule
& \multicolumn{3}{c}{\bf Overall} & \multicolumn{1}{c}{\bf Chest} & \multicolumn{1}{c}{\bf Hip} & \multicolumn{1}{c}{\bf Leg Length} & \multicolumn{1}{c}{\bf Waist} \\
\midrule
{\bf }  & TP90 & TP75 & TP50 & MAE & MAE & MAE & MAE \\ 
\hline
Single-View (No Aug.) & 23.32 & 15.43 & 8.74 & 22.74 & 16.64 & 13.72 & 20.88 \\
Single-View (Adv. Aug.) & \textbf{23.24} & \textbf{15.43} & \textbf{8.55} & \textbf{22.67} & \textbf{16.41} & \textbf{13.58} & \textbf{20.78} \\
\bottomrule
\end{tabular}
}
\caption{Comparison of No Augmentation~(No Aug.) versus Adversarial Augmentation for BMnet on the minimal clothing subdivision of TestB~(errors in mm).}
\label{table:supp_testb_minimal}
\end{table*}


\begin{table}[t]
    \centering
    \resizebox{0.9\columnwidth}{!}{%
    \begin{tabular}{lccc}
    \toprule
    {\bf } & Ours Adv. Aug. & Ours No Aug. \\ 
    \midrule
    Ankle & 5.56 & \textbf{5.48} \\
    Arm Length & \textbf{7.07} & 7.26 \\
    Bicep & \textbf{6.36} & 6.50 \\
    Calf & \textbf{7.89} & 7.95 \\
    Chest & \textbf{18.84} & 19.18 \\
    Forearm & \textbf{5.18} & 5.36 \\
    Head-to-Heel & \textbf{8.89} & 9.11 \\
    Hip & \textbf{11.34} & 11.53 \\
    Leg-Length & \textbf{11.24} & 11.39 \\
    Shoulder-Breadth & 6.05 & \textbf{5.95} \\
    Shoulder-to-Crotch & 8.90 & \textbf{8.85} \\
    Thigh & \textbf{11.15} & 11.16 \\
    Waist & \textbf{15.78} & 16.12 \\
    Wrist & \textbf{4.31} & 4.35 \\
    \midrule
    Mean Error & \textbf{9.19} & 9.30 \\
    \bottomrule
    \end{tabular}
    }
    \caption{Mean measurement error comparison of Adversarial Augmentation~(Adv. Aug.) versus No Augmentation (No Aug.) for training BMnet on TestA~(errors in mm).}
    \label{table:supp_measurements_normal}
\end{table}

\subsection{Limitations and Future Work } 
Similar to other adversarial training techniques, our method incurs a, small but real (order of 1\%), computational overhead to achieve improved accuracy. Techniques such as “Adversarial Training for Free!”  \cite{shafahi2019adversarial} may be explored to reduce training time and data storage. Our adversarial synthesizer currently does not account for environmental variations that affect the input silhouettes, such as camera characteristics, human pose variations, and segmentation noise. Adversarial sampling incorporating these dimensions is a fruitful future investigation.
While some methods perform test-time optimization~\cite{Joo2021}, the focus is usually on pose optimization rather than shape. Further improvement of all comparative methods in our work using test time optimization is interesting, but beyond the scope of this work.

\subsection{Societal Impact} 
1. Our system predicts intimate attributes about a person (i.e. body measurements) from photo silhouettes. These attributes are considered confidential, as they can be linked to one's health, personal lifestyle, and choices. It is therefore important that such a pipeline is protected from access by unqualified authorities who could generate and misuse confidential body information.
2. Computer vision research in the fashion domain has been supported by datasets that are heavily biased to thin and tall body shapes. This is owed to the preponderance of photos of models and celebrities from which these datasets are sourced~\cite{Liu_2016_CVPR, rostamzadeh2018fashiongen, 6248101}. Consequently, networks that estimate body shape and measurements, and generate body avatars for virtual try-on experiences, tend to produce larger errors for body shapes that deviate from societal beauty standards. Our research aims to increase inclusivity in body shape by discovering body shapes that are rare with respect to available datasets. However, a purely computational approach to countering dataset bias may also introduce other unfavorable biases; hence it is important to check for alignment between the body shape distributions generated by our method and realistic shape distributions in a given demographic. In our work, we attempt to address this issue by performing adversarial perturbations around real body shapes in BodyM.
\subsection{Ethics Statement} 
While we present research and datasets on human body measurement estimation, we take all precautions to respect the privacy of all individuals who have contributed to our data and research. Our collected human body dataset comprises silhouettes, height, weight and body measurements which do not reveal subject identity. The outputs of our adversarial body simulator are synthetic. All subjects have given written consent for the capture and release of the data.

\subsection{Reproducibility}
The BodyM dataset is publicly available at \url{https://adversarialbodysim.github.io} to enable reproducibility of our method and further research in this area.

